\documentclass[letterpaper, 10 pt, conference]{IEEEtran}
\IEEEoverridecommandlockouts
\usepackage{cite}
\usepackage{amsmath,amssymb,amsfonts}
\usepackage{algorithmic}
\usepackage{graphicx}
\usepackage{textcomp}
\usepackage{xcolor}
\usepackage{tikz}
\usepackage{hyperref}
\usepackage{rotating, multirow}
\def\BibTeX{{\rm B\kern-.05em{\sc i\kern-.025em b}\kern-.08em
    T\kern-.1667em\lower.7ex\hbox{E}\kern-.125emX}}

\newcommand\copyrighttext{%
  \footnotesize \textcopyright 2023 IEEE. Personal use of this material is permitted.
  Permission from IEEE must be obtained for all other uses, in any current or future
  media, including reprinting/republishing this material for advertising or promotional
  purposes, creating new collective works, for resale or redistribution to servers or
  lists, or reuse of any copyrighted component of this work in other works.}
\newcommand\copyrightnotice{%
\begin{tikzpicture}[remember picture,overlay]
\node[anchor=south,yshift=10pt] at (current page.south) {\fbox{\parbox{\dimexpr\textwidth-\fboxsep-\fboxrule\relax}{\copyrighttext}}};
\end{tikzpicture}%
}

\begin{document}

\title{What Could a Social Mediator Robot Do? 

Lessons from Real-World Mediation Scenarios}

\author{Thomas H. Weisswange$^{1}$, Hifza Javed$^{2}$, Manuel Dietrich$^{1}$, Tuan Vu Pham$^{1,3}$, \\Maria Teresa Parreira$^{4}$, Michael Sack$^{4}$, and Nawid Jamali$^{2}$

\thanks{$^{1}$Honda Research Institute Europe GmbH {\tt\smallskip \{thomas.weisswange, manuel.dietrich\}@honda-ri.de}, $^{2}$Honda Research Institute USA, Inc. {\tt\smallskip \{hifza\_javed, njamali\}@honda-ri.com}, $^{3}$University of Siegen, Siegen, Germany {\tt\smallskip tuan2.pham@uni-siegen.de}, $^{4}$Cornell University, Ithaca, NY, USA {\tt\smallskip \{mb2554, mjs596\}@cornell.edu}}

} %/author closing braces

\maketitle
\copyrightnotice
\newcommand{\delete}[1]{\textcolor{red}{#1}}

\begin{abstract}
The use of social robots as instruments for social mediation has been gaining traction in the field of Human-Robot Interaction (HRI).
So far, the design of such robots and their behaviors is often driven by technological platforms and experimental setups in controlled laboratory environments.
To address complex social relationships in the real world, it is crucial to consider the actual needs and consequences of the situations found therein. 
This includes understanding when a mediator is necessary, what specific role such a robot could play, and how it moderates human social dynamics.
In this paper, we discuss six relevant roles for robotic mediators that we identified by investigating a collection of videos showing realistic group situations.
We further discuss mediation behaviors and target measures to evaluate the success of such interventions.
We hope that our findings can inspire future research on robot-assisted social mediation by highlighting a wider set of mediation applications than those found in prior studies.
Specifically, we aim to inform the categorization and selection of interaction scenarios that reflect real situations, where a mediation robot can have a positive and meaningful impact on group dynamics.
\end{abstract}

\begin{IEEEkeywords}
human-robot interaction, social robotics, embodied mediation, group dynamics
\end{IEEEkeywords}

\section{Introduction}
\label{sec:introduction}

The traditional perspective on mediation is associated with the resolution of conflict with the help of an independent party. 
In the context of social relationships, it refers to a process for creating and repairing social bonds with an impartial party that seeks to help individuals or institutions to improve their relationship by organizing exchanges between them~\cite{jokinen2014social}.
Social mediation practice is commonly applied in workplace environments as a form of dispute resolution, 
but also
in the business-oriented role of meeting facilitators who support a group in achieving common goals~\cite{Viller1991facilitator}. 
Mediation is also applied in multicultural learning environments to cater to the social, cultural, and linguistic heterogeneity of the learners by providing diverse and more inclusive teaching practices and tools~\cite{de2005social, cesar2005curriculum}.

A relatively new application is the use of technological tools to serve as social mediators, given their ability to promote social relationships, groups, and communities~\cite{sutcliffe2011social} through social media. 
More recently, the use of robots for social assistance has started to gain traction. 
Such robots have been applied to a number of domains, including robots as companions~\cite{park2019model, leite2011modelling}, robots as tutors for children~\cite{gordon2016affective, park2019model}, and robots as assistants for older adults~\cite{broekens2009assistive, pedersen2018developing}, although in many cases through interaction with individual users only~\cite{Aylett2023}. 
In a group setting robots were tasked with direct actions to facilitate social interactions between human users or to equip humans with the skills to indirectly help them engage more effectively in social settings~\cite{palestra2016multimodal, marti2009creative, yee2012developing, lehmann2011make, pliasa2019can}. 
In this paper, we highlight a perspective on social robotic mediation that emphasizes the explicit improvement of social quality or interpersonal interactions, such as promoting user satisfaction, improving group cohesion, and fostering ingroup identification~\cite{abrams2020c}. 
This is distinct from improvements in a group's task performance, which may not necessarily lead to improved social interactions among group members~\cite{Short2017moderation}. 

Some existing approaches to achieving this goal using social mediator robots have been recently reviewed.
Abrams et al.~\cite{abrams2020c} proposed a theoretical framework reviewing key concepts from social sciences related to ingroup identification, cohesion, and entitativity, and proposed methods to measure these phenomena in robot-mediated human-human interactions.
Sebo et al.~\cite{Sebo2020} conducted a review of the literature on physically embodied robots that study group-level phenomena in human-human interactions, and addressed questions related to the effects of robot actions on group behaviors. 
Recently, Javed et al.~\cite{Javed2023} categorized existing approaches for robot-assisted social mediation into models of group dynamics and analyzed their ability to capture the relational aspects of human-human interactions. 
As apparent from these reviews, the approach to robot-based mediation has traditionally been technology-driven, where existing platforms and behaviors are evaluated in controlled settings involving groups of recruited participants. 
Although this approach has yielded interesting results, its applicability to real-world situations may be limited, since designing specific technical solutions for controlled experimental settings may 
constrain the effectiveness of such an approach in the real world.

In this work, we draw inspiration directly from human mediation behaviors found in existing everyday, real world scenarios to propose a novel situation-centric approach for robot-assisted social mediation design. 
We do so by collecting and reviewing video clips containing human-human interaction scenarios that either involve a human mediator or that may benefit from mediation. 
This paper makes the following contributions: first, it expands the understanding of social mediator robots by exploring new mediation roles and identifying specific actions a robot can produce to be effective in each role.
Second, our work identifies factors of social context that characterize the situations requiring mediation, which may be used to identify the mediation role a robot must play.
Third, we also identify measures to evaluate the success of mediation in each role, with an exclusive focus on social targets of mediation that can often be more challenging to determine than task-specific targets. 

\section{Video Collection and Analysis}
\label{sec:videos}
We wanted to gather a range of situations consisting of human group interactions that either involve mediation or may benefit from mediation. Our goal was not to identify every possible mediation situation, but rather to explore compelling categories of scenarios that could present prospects for future applications of robotic mediation.

We collected video material from publicly available online video-sharing platforms (no personal information or data of any individuals was collected). 
The authors of this paper were instructed to search for videos that could be used to stimulate discussions related to social mediation. 
Specifically, the videos were required to show a group situation either with a change in some group quality (e.g., starting conflict, change in activity level, etc.) or an active mediation process that could be understood from an approximately 30- to 60-second video excerpt.

To ensure diversity of the collected material, we did not specify search terms, but relied on each author's judgment when selecting videos that were relevant to the research question and met the inclusion criteria. The final set of 25 collected videos included situations from movies and real-world settings, including exaggerated, rare, but also everyday interactions, with group sizes ranging from 2 to greater than 30. During a two-day in-person workshop, we analyzed the videos and discussed the possible classification and description of different mediation roles.

\section{Robot-Assisted Social Mediation Roles}
\label{sec:roles}
\begin{figure}[!ht]
\centering
\includegraphics[width=\columnwidth]{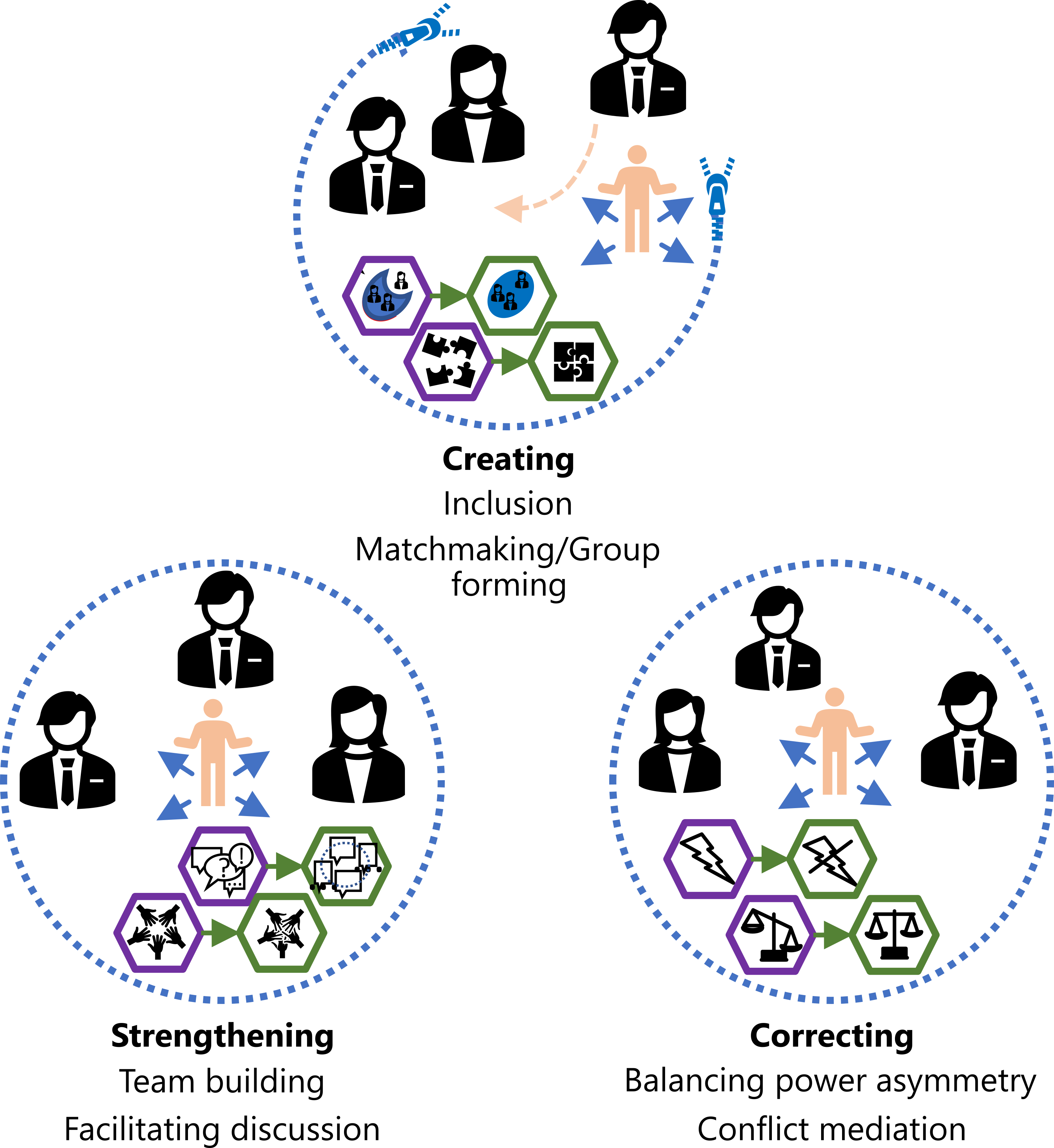}
\caption{Roles for social mediation}
\label{fig:roles}
\end{figure}

We derived six frequently encountered themes for social mediation:
A) facilitating discussions, B) resolving conflicts, C) team-building and coaching, D) inclusion, E) group formation and matchmaking, and F) balancing power asymmetry (Fig.~\ref{fig:roles}). 

In the following section, we 
reflect on their characteristic contextual factors that 
are known to have a strong influence on the socio-emotional states of interactants \cite{greenaway2018context, mesquita2014emotions} (such as personal, relational, situational, and cultural context). 
For each category, we also highlight the potential measures that can be used to track and evaluate the impact of mediation, identify the specific actions used by the mediator to achieve the mediation goal (see Table \ref{tab:mediation-types} for an overview), and relate to existing approaches that may have explored such a role.

\begin{table}[!htbp]
    \caption{Types of mediation behavior}
    \label{tab:mediation-types}
    \begin{center}
    \begin{tabular}{|p{0.13\columnwidth}|p{0.15\columnwidth}|p{0.57\columnwidth}|}
        \hline
        & Category        & Description \\
        \hline
        Interact & Rewarding  & Enhance positive behavior            \\ 
        & Correcting          & Improve non-positive behavior        \\ 
        & Interrupting        & Stop negative behavior             \\ 
        & Connecting          & Improve relations between individuals  \\ 
        \hline
        Inform & Structuring  & Enforce rules or schedules          \\ 
        & Motivating          & Emphasise group identity or targets \\
        & Grounding           & Create common understanding         \\
        & Awareness-Raising   & Stress impact of ongoing group dynamics  \\ 
        \hline
        Influence & Leading   & Display positive behavior for imitation   \\ 
        & Nudging             & Use subliminal or peripheral communication \\ 
        & Atmosphere-Creating & Change environmental influence factors     \\ 
        \hline
    \end{tabular}
    \end{center}
\end{table}

\subsection{Facilitating Discussions}
\label{subsec:facilitating}
The first setting involves groups engaged in discussions to reach a mutual goal and a mediator that tries to improve the involvement of and interactions between the group members as they attempt to reach this goal.
Typical tasks require the group to find an agreement on a decision between different options or to create new ideas and solutions to a given problem. 

\subsubsection{Contextual factors}
Situational factors describing such a setting typically include
an indoor setting with a co-located, small to medium-sized group (often 3--8 people). 
In many cases, this will be a classical ``meeting room'' environment with group members sitting around a table.
The participants often share a business-related background, although not necessarily implying uniform skills or knowledge bases.
A need for mediation will most prominently arise from contextual factors such as groups with differing opinions or preferences, a mix of different personalities, potential meeting role implications, as well as inter-individual relationships and interaction histories.
In contrast to these aspects, a shared goal or decision theme will often be established already.

\subsubsection{Possible measures}
In the case of group discussions, a major mediation goal may be to achieve positive affective states for all group members. 
Previous work proposed methods to measure this both continuously during the discussion or at the end of a meeting \cite{sharma2019automatic, smith2015real, vonikakis2016group}.
The participation of individual members in the discussion may be measured through explicit contributions but also through back-channeling behaviors \cite{Coker1987involvement, tennent2019micbot} that can positively influence in-group identification.
Apart from participation, measuring the balance in the utilization of input between all members, for example through speaking times or overlaps in speaking activity, can also be used as an indicator of the group's cohesion in general~\cite{hung2010estimating}.
There exist also measures for evaluating the decision or creativity output of a group (e.g. \cite{de2003task})
However, for the social qualities, it might be sufficient to guarantee that there is an outcome that the members are satisfied with, which may, for example, be achieved through the use of sub-scales of the Subjective Value Inventory~\cite{curhan2006people}.

\subsubsection{Mediation behaviors}
When looking at the behaviors of human mediators in group discussions, we found both individual-directed and group-directed behaviors. 
Most behaviors could be framed as strengthening the group dynamics, for example through appraisal of contributions (of specific members), independently of its content. 
This was done through verbal interactions, such as saying ``thank you'', or non-verbal behaviors, such as gaze and smile or through a pat on the back.
Examples of group-directed rewarding actions are descriptions of the positive outcomes of the discussion quality, but also more subtle behaviors such as a relaxed or interested pose.
We also observed structuring interventions that tried to guide conversations between only a subset of the group or about an off-topic back towards a focused and balanced discussion.
We find explicit verbal statements that raise awareness of problematic dynamics or hint at how to improve the dynamics, as well as, simple utterances that try to break an undesired dynamics.
In creativity-oriented meetings, human mediators also try to disrupt stalled discussions through provocative or silly questions or by changing the seating order.

\subsubsection{Existing robotic approaches}
Many existing robotic mediation studies aim at some version of meeting facilitation.
The Micbot \cite{tennent2019micbot} tried to mediate group engagement through enhancing a good distribution of speaking times in a group problem solving context.
In \cite{Bohus2010facilitating, Parreira2022}, a robot tried to balance participation through gaze behaviors.
\cite{Shamekhi2019multimodal} constructed a robotic system to facilitate meetings with respect to structure and task progress, but also included behaviors to encourage the participation of members with low speaking time. 
However, in many studies, the focus is on improving the group's output success rather than the social dynamics, which may be partially attributed to using a group of strangers in an artificial decision making setting.
An interesting exception was described by Erel et al. \cite{Erel2021hhri}, who introduce a robotic object that was shown to improve the mutual emotional evaluation of a dyadic conversation through mediation with simple body movements.
We would like to encourage future work to include estimation of group states through a larger set of multi-modal behaviors, scalability of group state determination and mediation methods to larger group sizes, and the use of more sophisticated and nuanced mediation actions that can influence existing social dynamics between the interactants.

\subsection{Resolving Conflicts}
\label{subsec:conflict}
Group interactions can lead to conflicts between members, which,  if not resolved, may impact the quality of the interpersonal dynamics.
Two types of conflicts can occur within groups: relationship and task conflicts~\cite{jehn1995multimethod} and both types of conflict can also negatively impact a group's performance~\cite{de2003task}. 
Mediation may support conflict resolution in both short- and long-term. 
In fact, the latter is the setting to which human mediators are classically related.

\subsubsection{Contextual factors}

Various personal factors, including personality traits, emotional states, and individual values and goals, can significantly impact the way a conflict unfolds and the strategies used to resolve it.
Relational context might be particularly useful in predicting or preventing conflicts, where, for example, prior interaction history or power imbalances could provide 
useful insights that a mediator may use to adjust their intervention.
Whether a conflict involves mental, verbal, or physical altercations may also impact the means necessary for mediation.
Similarly, cultural context may influence the risk of misunderstandings, and existing sub-group dynamics can determine if the conflict is at risk to grow beyond the originally affected parties.

\subsubsection{Possible measures}
A conventional approach to evaluate conflict management may directly measure the affect of the persons involved in the conflict. 
However, we believe it is also important to follow the emotional impact of the mediation on the other group members, for example through measuring group affective balance \cite{jung2016coupling}.
Long- and short-term effects on group membership can be evaluated by using group cohesion or ingroup identification scales \cite{brawley1985development, leach2008ingroup}.
There also exist conflict severity measures (e.g. \cite{danes2000effects, Kim2012conflict}) that could be used to follow the influence of mediation behaviors on conflicts with slower dynamics.
As increasing cooperation within the group correlates with positive conflict resolution \cite{lewicki1992models}, cooperation-related measures (e.g. \cite{Manson2013coop, Wollstadt2022}) may also be useful for monitoring and evaluating the success of a mediation.

\subsubsection{Mediation behaviors}
Many of the mediation behaviors that we found in the videos were direct interventions that served to interrupt the conflict behaviors, either verbally or physically.
For example, positioning one's body between the involved individuals to impair eye contact was successful to calm down heated debates.
At an early stage of a conflict, it may be sufficient to call out an aggressive or provocative action from any interactant, prompt interactants to take a break from an argument, or involve other group members that may provide additional information or compromises.
Moderation behaviors that aim to structure an interaction by, for example, highlighting social rules, or pointing to alternative approaches to address a given task, can also help prevent or even resolve conflicts.
Playing music, changing the lighting, or telling a joke may lighten the group ambience and reduce negative emotions that can cause conflicts among participants.
In preparation for a group event, a mediator can also try to arrange the seating order to prevent individuals who are prone to conflict from sitting together.
Interestingly, we also found cases where a mediator allowed the conflict to unfold, only preventing bystander involvement, and eventually coordinating a debriefing that helped improve overall group dynamics.

\subsubsection{Existing robotic approaches}
Druckman and colleagues \cite{druckman2021best} compared a (teleoperated) robot to a human and a screen-based solution for mediating a company de-merger meeting. 
Besides positive impact on the number and type of agreements, robotic negotiators caused higher participant satisfaction with the meeting outcome.
In \cite{shen2018stop}, a robot mediated object possession conflicts between children, and in \cite{jung2015using}, a robot intervened in a team-based problem-solving task by repairing a task-directed or personal attack from a confederate. 
The researchers found that such repair interventions heightened awareness of a normative violation.
We encourage future work to include strategies targeting conflict prevention, in addition to methods of conflict resolution. 
Determining the optimal timing of an intervention is also an interesting problem to address in the future, given that some conflicts may be best resolved by the involved individuals, while others may depend on external intervention, in which case it would be interesting to see robot behavior strategies involving bystanders.

\subsection{Team Building and Coaching}
This category of mediation scenarios covers roles in which the mediator works with a group that shares a common long-term goal (a "team"), or is bound by social ties, such as in a family. 
A major difference from the previous categories is the focus on strengthening long-term group dynamics. 
Interventions are often performed before (or sometimes after) a group engages in a concrete event.
Another interesting aspect of this role is the support of the transformation of a short-term group into a long-term team.

\subsubsection{Contextual factors}
A mediator usually works with groups having high entitativity \cite{campbell1958entitativity}, which means that a bystander is easily able to identify members of that group.
The group's common goals are likely already well-established.
Similarly, certain interests, personal background, or skill aspects will often be shared, as a major defining feature of a team. 
However, other dimensions, such as cultural or social backgrounds, may vary greatly among the members. 
For example, members of a football team will generally have above-average sports skills, but may come from diverse cultural or social backgrounds. 
Over time, different roles may be established within the group, and these roles will influence the dynamics of the team. 
Due to the stability of the group, there will likely be relevant historical interactions among group members, in addition to shared memories.

\subsubsection{Possible measures}
Since teams usually exist for longer periods, the success of this mediation role is primarily evaluated using long-term measures. 
This may make it challenging to immediately assess the quality of an intervention, but it enables the use of questionnaires or interviews for feedback, whereas other roles are bound to interpret sensor data for timely evaluations. 
For teams, two essential social qualities are mutual trust and a high level of shared understanding.
There exists different ways to measure trust in teams depending on the context~\cite{McEvily2011trust} and also several methods to assess shared understanding, including a construct called team mental models~\cite{Klimoski2016}.
In general, mediation also aims to enhance existing group qualities, including in-group identification and cohesion.

\subsubsection{Mediation behaviors}
Mediators use speeches to motivate teams and enhance their group qualities. 
This is done explicitly, such as by highlighting shared backgrounds, or implicitly, through referencing shared knowledge or making in-jokes based on past experiences. 
They may also highlight differences with other groups. 
Some mediation actions target enhancing physical closeness between the team members, such as by gathering the team for a huddle or wrapping arms around multiple members. 
To strengthen mutual relationships, mediators tell anecdotes that explicitly name two or more members. 
To make individuals feel like important parts of the team, we saw gesture towards them and behaviors that explicitly highlighted their roles within the team. 

\subsubsection{Existing robotic approaches}
There appears to be little existing research on robots for team-building. 
In \cite{Utami2017couples}, a social robot-assisted in counseling couples (a minimal team) by teaching and demonstrating communication methods that could enhance long-term relationships. 
In a paper by Stoican et al.~\cite{Stoican2022}, a concept was proposed for mediating trust-building between humans. 
Short et al.~\cite{Short2017moderation} used a social robot to support a group in playing a collaborative game, where it provided task-based recommendations, either to optimize performance or participation balance.
It is interesting to note that the group using the performance-optimizing mediator had better immediate and post-intervention performance, while the other group showed an improvement in group cohesion.
Overall, there are still many open issues around team-building with social mediator robots. In particular, considering long-term effects and using already established groups in experimental settings are two aspects that we hope future work will address. 

\subsection{Inclusion}
Mediation can be highly effective in integrating new members into an existing group, such as when a new employee is hired in a business setting or a family moves to a new neighborhood.
This category also includes situations within an existing group where individual participants feel isolated or pushed to the fringes due to ongoing dynamics, discussions, or task context. 
This can happen unintentionally and even without the awareness of other group members.
If left unaddressed, it can weaken the group's social dynamics to the point where a member may consider leaving.

\subsubsection{Contextual factors}
This role is particularly relevant in groups with higher fluctuations or weaker organizational or socio-emotional coupling, such as volunteer organizations, book clubs, senior citizen clubs, recreational activities like fitness classes, gardening or art classes, and social events where people are mingling. 
These groups are often drawn together by a single shared interest, but may face challenges in providing a welcoming environment to all members.
In a business setting, a shared task or knowledge base may provide motivation for inclusion, but in casual or social situations, newcomers may feel insecure or timid to approach a group, and existing members may not see an immediate benefit.

\subsubsection{Possible measures}
To evaluate the success of mediation in this role, a combination of measures proposed for previous categories can be used. 
The most straightforward measure may be the ingroup identification of the new or left-out member. 
Cohesion can be used to assess the impact of including the new member on the group's qualities. 
As a continuous measure, the engagement of the new member, the balance of contributions between new and old members, or the proxemics of the group~\cite{Rosatelli2019} may be evaluated. Additionally, measures of interaction strength and collaboration, which are covered in entitativity~\cite{campbell1958entitativity, Lickel2000}, may be made accessible to a robot to help evaluate the success of its behaviors.

\subsubsection{Mediation behaviors}
One set of mediation behaviors aims at establishing the group context. 
This may entail using verbal prompts to connect individuals based on the group's central theme, or physical interventions that modify positioning, such as guiding the person in question towards the center of the space or increasing overall clustering. 
Another strategy aims to reduce the barriers for a newcomer to integrate into the group. 
A mediator may verbally encourage others to join, move closer, or contribute to the main group through the use of positive emotional expressions, gestures, or verbal cues. 
They may also redirect the conversation to a more casual topic or provide a clear task to help alleviate feelings of vulnerability. 
Similar behaviors can also be beneficial when a group member feels excluded, such as when they cannot contribute to the current topic or task. 
Sometimes, simply raising awareness is enough to promote re-inclusion into the group---for instance, a mediator may gaze or orient their body towards that person.

\subsubsection{Existing robotic approaches}
Matsuyama et al.~\cite{Matsuyama2015} proposed a conversational robot that attempts to include a person that may be feeling excluded from a conversation. 
It does so by getting involved in the conversation and redirecting attention towards the excluded participant.
Mutlu et al.~\cite{Mutlu2009footing} demonstrated that a robot can use gaze to establish a participant as addressee or bystander in a group conversation.
A robot mediator trying to support the integration of immigrant children into existing groups by promoting participation was presented in \cite{Gillet2020}.
Another study \cite{neto2023robot} used a robotic mediator to support inclusion between children with different levels of visual impairment by assessing the effect on speaking time distribution.
A robot that shows vulnerability through verbal statements was shown to positively influence the conversation volume~\cite{traeger2020vulnerable}.
However, Sebo et al.~\cite{Sebo2020inclusion} showed that a robot might not always help in mediating the inclusion of outgroup individuals if its action is perceived as favoring one side.
Current work seems to rely heavily on the use of verbal activity to detect outgroup situations. In the future, we would recommend the use of additional modalities, for example, visual cues such as interpersonal distance and gestures, to obtain a more comprehensive understanding of such group dynamics and design more effective interventions to promote inclusivity.

\subsection{Group Formation and Matchmaking} \label{matchmaking}
This category covers situations in which groups are not established yet. 
Its two main aspects are: 1) to help build new connections between people by matching individuals that have a high chance of a future social relationship with each other, and 2) to form a group out of individuals.
A conventional application of social matchmaking systems is online recommender systems for people, with a strong focus on romantic relationships and interest groups in social networks~\cite{Terveen2005}. 
However, the need for such mediation exists within in-person social interactions as well.
An example of this may be larger gatherings, such as weddings or professional networking events, where people may often be unfamiliar with each other.

\subsubsection{Contextual factors}
For situations relevant to this category, the most important ability in a mediator is to detect the opportunity for group formation.
This may be achieved by leveraging commonalities in individuals' backgrounds, interests, and/or other relevant factors, or by considering tasks that may be beneficial for individuals to undertake in groups.
A typical contextual feature common to such situations can be the lack of familiarity between the individuals and lack of knowledge of existing commonalities, which may often manifest in hesitation in deciding who to approach and in making the first contact once this decision is made.

\subsubsection{Possible measures}
In matchmaking opportunities, a mediator can often receive immediate feedback, as people may simply reject to initiate contact at all or cut interactions short.
If the first contact is successful, measures of general joint activity, involvement of all members, and balance of contributions can provide meaningful insights.
Some form of ingroup identification may help identify the validity of the matchmaking strategy if it is not possible to follow a new group for a sustained duration.
Otherwise, entitativity and cohesion can be evaluated after an extended interaction.

\subsubsection{Mediation behaviors}
Common behaviors for matchmaking mediators aim to co-locate the individuals, either by organizing an event or by setting up a specific seating order, or by physically guiding the matched individuals to a common space.
To provide a common ground, it can be useful to make explicit introductions by highlighting shared interests or providing an environment that implies certain commonalities.
To overcome possible hesitation at the start of an interaction, a mediator may provide a starter question or activity, and promote a relaxed atmosphere by telling a joke.  

\subsubsection{Existing robotic approaches}
The technology for matching people based on personal data has existed for some time now~\cite{Terveen2005}
but matchmaking has been mostly restricted to computer interactions. 
An approach to mediation of face-to-face meetings used a smart device for the detection of mutual availability to suggest and initiate a video connection \cite{Einecke2022}.
A study \cite{Terry2002SocialNet} proposed to trigger human users to act as matchmakers if their network contained mutually unknown friends that frequently visit the same locations. 
Xu et al. \cite{Xu2014SoBot} proposed a mobile phone app that generated mutual text introductions at a first meeting. 
A chatbot proposed topics of shared interests~\cite{Shin2021}, while a social robot provided ``icebreaker'' topics \cite{Zhang2023icebreaking}.
An early conceptual paper described multiple mobile robot companions that would guide their users to incidental meetings with others with similar interests \cite{Ono1999}. 
Overall, there are only a few existing approaches in this area, which might make it an interesting direction for future research.

\subsection{Balancing Power Asymmetry}
In the reviewed videos, we also identified situations in which the group members held different hierarchical positions or had other means of implicit or explicit power over others.
In general, social power is the existence of a resource or emotional dependence of one or multiple people on another \cite{emerson1962power}.
Such situations can, for example, lead to low-power members not contributing or opportunistically following the opinions of the high-power members.
In general, the group dynamics tend to drift towards states with high satisfaction or positive affect of few, powerful, members, often at the expense of the others.
Power asymmetry can facilitate negative behaviors with possible long-term effects on group and individuals, such as displays of dominance, discrimination, and favoritism, although high-power individuals might also benefit a group~\cite{larson1998leadership}.
One target for a mediator can be to change a power imbalance when an interaction can benefit from diluting or breaking down the hierarchical boundaries.

\subsubsection{Contextual factors}
Power asymmetries commonly arise in work or school contexts, where group members' personal aspects, such as education, gender, age, or economic status, may result in implicit power asymmetries. 
Explicit role assignments, such as those between managers and workers or teachers and students, can also create momentary power imbalances within a group.

\subsubsection{Possible measures}
Social dominance in interaction can be measured through its correlation with body movements of individuals \cite{escalera2010automatic, jayagopi2009modeling} or specific conversation patterns (speaking time, speech volume, tempo, pitch, vocal control)~\cite{dunbar2005perceptions}.
Another interesting measure might be cross-understanding~\cite{Huber2017crossunderstanding}, which evaluates the mutual understanding of each other’s mental representation of the situation, which can address misinterpretations of power distributions and power utilization.

\subsubsection{Mediation behaviors}
Mediators prepare an environment in a way to prevent awareness of power asymmetries in groups, for example, through strategic seating arrangements.
We observed instances where a desired disregard for hierarchies was communicated explicitly or through actions such as playing a lighthearted game before initiating the intended interaction.
Establishing social rules can also provide a guideline of where dependencies might be ignored safely. 
Since a mediator may be perceived as neutral and/or independent, it might act as a proxy of low-power members' opinions or point out dominance behaviors without raising bad feelings for the higher-power individuals.
Dominance behavior may also be interrupted through physical interventions or simple distracting utterances.

\subsubsection{Existing robotic approaches}
Skantze~\cite{skantze2017predicting} presented a conversational robot that could reduce imbalance in contributions by explicitly addressing the least dominant member, and also proposed measures to predict power asymmetries early during an interaction. 
In a similar approach, robot gaze was used to improve participation of non-dominant group members~\cite{gillet2021robot}. 
Mediating dominance behavior is often handled as a part of meeting facilitation. 
However, we believe that it is worth considering on its own since it includes specific mediation behaviors and targets involving long-term group qualities.

\section{Discussion and conclusion}
\label{sec:discussion-conclusion}
In this paper, we provided a new perspective on possible roles for social robots for mediation of groups of humans in alternative to the dominating technology-centered approach. 
We identified mediation roles in Section~\ref{sec:roles}, covering three different impact forms of a mediator on the dynamics of a group (Fig. ~\ref{fig:roles}). 
In a situation where positive group dynamics exist between the members, a mediator would strengthen and further enhance these. 
If a group shows flaws in its dynamics, a mediator would take corrective actions to address the shortcomings.  
Finally, in situations where pre-existing groups are absent or incomplete, the mediation target would be to enable the creation of dynamics between the present individuals.
Although, there exists prior work on robotic mediation, many approaches are focused on meeting facilitation and some aspects of conflict resolution. 
We acknowledge the potential difficulties in setting up laboratory experiments for the other roles
in particular when it comes to studying long term effects. 
However, we would like to encourage researchers to consider these roles in the design of next generation robot mediators, since they
are important for a robot's real world relevance and positive social impact. 

We described the most prominent mediation scenarios identified through our video analysis activity, 
however, we also encountered additional mediation scenarios (for example controlling crowd dynamics to prevent escalations) that we deemed not to be fully aligned with the theme of this paper. 
Interestingly, we also found situations where a mediator used information about individual group members to motivate a team while avoiding violation of an individual's privacy. 
Privacy-related concerns around social mediator robots have been explored in~\cite{Dietrich2022privacy}.
It must be noted that analyzing mediation behaviors based on human examples may carry the risk of
biasing robot designs towards acting in the role of the human, while there also exists support to investigate potentially unique robotic ``superpowers''~\cite{Dorrenbacher2023superpowers, Welge2016superpowers}.
However, for this work, our goal was not to provide a comprehensive definition of robot mediators but rather to offer intriguing insights for future research that can expand the scope of the field to include real-world scenarios.

%\section{Acknowledgments}
%\label{sec:acknowledgements}
\bibliographystyle{IEEEtran}
\bibliography{main}

\end{document}